# The Quantum Challenge in Concept Theory and Natural Language Processing

Diederik Aerts[1], Jan Broekaert[1], Sandro Sozzo[1] and Tomas Veloz[1,2]

[1] Center Leo Apostel for Interdisciplinary Studies and, Department of Mathematics, Brussels Free University, Krijgskundestraat 33, 1160 Brussels, Belgium.
E-Mails: diraerts@vub.ac.be, broekae@vub.ac.be, ssozzo@vub.ac.be

[2] Department of Mathematics, University of British Columbia, 3333 University Way, Kelowna, British Columbia, Canada. E-Mail: tomas.veloz@ubc.ca

**Abstract**
The mathematical formalism of quantum theory has been successfully used in human cognition to model decision processes and to deliver representations of human knowledge. As such, quantum cognition inspired tools have improved technologies for Natural Language Processing and Information Retrieval. In this paper, we overview the quantum cognition approach developed in our Brussels team during the last two decades, specifically our identification of quantum structures in human concepts and language, and the modeling of data from psychological and corpus-text-based experiments. We discuss our quantum-theoretic framework for concepts and their conjunctions/disjunctions in a Fock-Hilbert space structure, adequately modeling a large amount of data collected on concept combinations. Inspired by this modeling, we put forward elements for a quantum contextual and meaning-based approach to information technologies in which 'entities of meaning' are inversely reconstructed from texts, which are considered as traces of these entities' states.

## Introduction - What is Quantum Cognition?

Cognitive phenomena reveal processes and properties akin to quantum systems. These notions correspond to *contextuality*, i.e. the non-deterministic change of state of the system that occurs whenever a measurement is performed on it, *entanglement*, i.e. the intrinsically correlated properties in composite systems, and *emergence*, i.e. the impossibility to describe a system purely by its components (Aerts et al., 2013b). While initially the mere identification of these structural properties was established in specific cognitive phenomena (Aerts & Aerts, 1995), later full quantum-based models were developed in a wide range of areas, e.g., different situations of human cognition in psychology and economics, finance systems, and ecological situations of population dynamics. For a comprehensive review we refer to the quantum cognition webpage *http://en.wikipedia.org/wiki/Quantum_cognition,* and references therein.

The quantum framework has also been intensely applied to information systems. This approach initiated with the application of quantum models in Information Retrieval (IR) (van Rijsbergen, 2004), and with the identification of quantum structures in existing semantic theories, such as LSA (Aerts & Czachor, 2004). Meanwhile, a multitude of approaches, that make use of the quantum notions were developed in models of word meaning (Widdows, 2006), contextual IR (Melucci, 2008), document ranking (Zuccon &

Azzopardi, 2010), etc. For a comprehensive review on quantum-models in information systems, see, e.g., Melucci & van Rijsbergen (2011).

In this paper we review the research in quantum cognition developed at the research center Leo Apostel (CLEA), led by Diederik Aerts, and propose a novel quantum-based and first principles interpretation of information systems.

**The Brussels Approach in Quantum Cognition**

Prior to the emergence of quantum cognition, quantum structures were identified in macroscopic systems. In Aerts (1982), the first example of a macroscopic system – communicating vessels – exposing measurements that violate classical probabilistic invariants (*Bell inequalities*), was demonstrated. The probabilities of such systems can exactly match the correlations of a quantum system consisting of a two spin-1/2 entities (Aerts, 1991). Subsequently, other macroscopic examples violating Bell inequalities were described, see Aerts et al. (2000) for a review. Regardless the nature of the system, the violation of Bell inequalities, and thus emergence of quantum probabilities, occurs when there is (a) a change of state of the system provoked by the measurement and, (b) a lack of knowledge about the measurement process (Aerts, 1986). Parallel to this, the exploration of quantum structure in opinion polls was initiated under the assumption that, in many cases, the opinion of a subject about a certain matter is interactively created, rather than being predefined (Aerts & Aerts, 1995). In this sense, the subject is under the influence of a *contextual landscape* given by its interaction with the questionnaire.

The analogy between quantum measurements and contextual landscapes was formalized later in the *state context property* (SCOP) theory of concepts (Aerts & Gabora, 2005). In SCOP, concepts refer to fundamental *meaning entities*. A concept exists in a potentiality state until a context is activated. This contextual interaction triggers a concept-collapse from a potential to an actual state of meaning. SCOP has modeled several context-dependent meaning estimations made in psychological studies (Gabora & Aerts, 2002).

It is remarkable that SCOP allows modeling the elusive problem of non-compositional meaning, discussed in many areas such as categorization, pragmatics, semiotics, etc. (Fodor, 2001). For example, it is known that the typicality (or other meaning estimation such as membership, similarity, etc.) of an exemplar, e.g., *Guppy*, with respect to a combined concept *Pet-Fish* cannot be explained in terms of the typicalities of the exemplar with respect to two component concepts *Pet* and *Fish* using classical or Fuzzy logics (Osherson & Smith, 1981). Particularly, a large amount of data collected in concept combination (Hampton, 1988a,b) cannot be modeled in a classical probabilistic setting (Aerts 2009). Our quantum-like representation of joint entities allows the existence of superposed states such that their exemplars' typicalities interfere with each other. This idea has been put forward in concrete Hilbert and Fock space-based models for concept combination of conjunctions/disjunctions (Aerts, 2009; Aerts et. al., 2012; Aerts & Sozzo, 2013a; Aerts, Gabora & Sozzo, S., 2013d).

A second fundamental area of research is the study of known paradoxes and effects in economics and decision theory, e.g., *Allais*, *Ellsberg* and *Machina paradoxes*, *conjunction fallacy*, *disjunction effect* (see, e.g., Allais, 1953; Ellsberg, 1961). These situations reveal a violation of the expected utility hypothesis and the Sure-Thing principle, basic assumptions in rational approaches to decision making under uncertainty (von Neumann and Morgenstern, 1944). Many approaches were propounded extending existing ones, but none of them was successful. In the Brussels approach, instead, a unified quantum framework was elaborated for both the disjunction effect and the Ellsberg and Machina paradox situations. In both cases, it was shown that it is the overall conceptual landscape which is responsible of the deviations from classical expectations typically met in these situations (Aerts et al., 2012; Aerts and Sozzo, 2013a).

**Towards a Theory of Meaning Representation**

The study of quantum structures in cognition, and the rigorous representation of these structures in Hilbert and Fock spaces, sustain the hypothesis that conceptual entities and quantum particles may respond to the same fundamental systemic principles, namely, contextuality, emergence and entanglement (Aerts, 2009; Aerts, 2010; Aerts and Sozzo, 2011; Aerts et al., 2012). The extent of this correspondence hypothesis is however still unknown. E.g., in quantum theory the state collapse occurs in 'space-time' along possible trajectory-traces, for conceptual entities however a corresponding realm for trajectory-traces and `localizing' collapse is not aptly defined. Recently, in Aerts et. al (2013c) it has been proposed that such realm could be circumscribed to be a corpus of text. Notably, pieces of text correspond to physical (or digital) representations of the meaning or ideas that the writer wants to convey. Therefore, our quantum-inspired framework assumes that (a) meaning is structured in the form of entities of meaning (concepts and their combinations) and, (b) the states of these entities of meaning collapse to a piece of text as a consequence of the interaction between the full non-collapsed states of the entity of meaning (concept) and the context contained in the piece of text, expressing this meaning within this specific piece of text. Therefore, a direct analogy between the study of quantum particles at the laboratory (e.g., with a spark chamber at CERN), and the study of meaning entities (concepts) from pieces of text can be made. Namely, quantum particles correspond to meaning entities, and the particle-trajectories observed at the laboratory correspond to the pieces of text in document sets (or chats, web page comments, blog entries, etc.).

To deepen the analogy above, a methodology to reconstruct the full non-collapsed states of such entities of meaning from their traces, i.e. their collapsed states, is needed. This procedure is known as the 'inverse problem' in mathematical physics, and it focuses on the existence, and eventual uniqueness of a system that delivers certain observed outcomes (Tarantola and Tarantola, 1987). Hence, our approach consists of reconstructing the full non-collapsed states of entities of meaning, structured as SCOP concepts, from textual data such as corpus of text or web pages.

This proposal involves many challenges, among which the need to scrutinize the extent to which SCOP functionally models entities of meaning and, a formalization of the notion of conceptual trace. The application of linguistic tools such as sentence processing (Hale

& Smolensky, 2004) can be applied to separate a piece of text into well defined conceptual traces. On the other hand, the underlying space where pieces of text are regarded as traces of concepts needs to be defined. Contrary to 'space-time' defined by physical bodies, the emergent topology of a 'space' for words and sentences remains unclear. This challenge hence, involves the development of a *semantic topology*. In this vein, some basic aspects have been noticed, such as the violation of metric axioms (Tversky, 1977), and formation of small world graphs (Choudhury & Mukherjee, 2009) by word-similarity data. However, in all approaches to develop a semantic topology, this was attempted without the distinction between collapsed and full non-collapsed states which in our approach is crucial. Hence, just like space only delivers a topology for collapsed states, we can focus on the set of collapsed states for the derivation of this topology. Hilbert and Fock type spaces, or more generally SCOP-type structures, generated over this topology in the collapsed region of the inverse problem, can lead to the identification of the overall structure. The above-sketched endeavor requires a collaborative effort in mathematics, physics, computer science and linguistics. We believe that this approach may lead to novel and possibly revolutionary technologies in the field of Natural Language Processing and Artificial Intelligence.